\documentclass{article}

\usepackage[preprint,nonatbib]{neurips_2023}

\usepackage[utf8]{inputenc} % allow utf-8 input
\usepackage[T1]{fontenc}    % use 8-bit T1 fonts
\usepackage{hyperref}       % hyperlinks
\usepackage{url}            % simple URL typesetting
\usepackage{booktabs}       % professional-quality tables
\usepackage{amsfonts}       % blackboard math symbols
\usepackage{nicefrac}       % compact symbols for 1/2, etc.
\usepackage{microtype}      % microtypography
\usepackage{amsmath,amssymb}
\usepackage{graphicx}
\usepackage{subcaption}
\usepackage{amsmath}
\usepackage{amsthm}
\usepackage{amssymb}
\usepackage{bm}
\usepackage{mathtools}

\usepackage[dvipsnames]{xcolor}
\usepackage{tabularx}
\usepackage{algorithm}
\usepackage[noend]{algpseudocode}
\usepackage{float}
\usepackage{wrapfig}
\usepackage{multicol}
\usepackage{multirow}
\usepackage{enumitem}
\usepackage{cite}  % You think it is OK to use this? for abreviating the reference lists

\newcommand\etc{etc\@ifnextchar.{}{.\@}}

\definecolor{CNN}{RGB}{66, 133, 244}
\definecolor{hybrid}{RGB}{84, 25, 85}
\definecolor{transformer}{RGB}{219, 68, 55}
\definecolor{baseline}{RGB}{116, 116, 116}
\definecolor{harmonized}{RGB}{236, 178, 46}
\definecolor{adversarial}{RGB}{127, 188, 0}
\definecolor{grey}{RGB}{143, 143, 143}

\newcommand{\pred}{\bm{f}_{\theta}}
\newcommand{\vx}{\bm{x}}
\newcommand{\vy}{\bm{y}}
\newcommand{\vp}{\bm{\phi}}
\newcommand{\gt}{\bm{\phi}}
\newcommand{\std}{\bm{z}}
\newcommand{\explainer}{\bm{g}}
\newcommand{\pyramid}{\mathcal{P}}

\theoremstyle{plain}

\theoremstyle{definition}

\theoremstyle{remark}

\title{Adversarial Alignment: breaking the trade-off between the strength of an attack and its relevance to human perception}

\author{%
  Drew Linsley$^{*1,2}$, Pinyuan Feng$^{*3}$, Thibaut Boissin$^{4}$, Alekh Karkada Ashok$^{1}$, \\ \textbf{Thomas Fel$^{5}$, Stephanie Olaiya$^{1}$, Thomas Serre$^{1,2,3,5}$} \\
  \texttt{drew\_linsley@brown.edu}
}

\begin{document}

\maketitle

\begin{abstract}
Deep neural networks (DNNs) are known to have a fundamental sensitivity to adversarial attacks, perturbations of the input that are imperceptible to humans yet powerful enough to change the visual decision of a model~\cite{Szegedy2014-qj}. Adversarial attacks have long been considered the ``Achilles' heel'' of deep learning, which may eventually force a shift in modeling paradigms. Nevertheless, the formidable capabilities of modern large-scale DNNs have somewhat eclipsed these early concerns. Do adversarial attacks continue to pose a threat to DNNs?

In this study, we investigate how the robustness of DNNs to adversarial attacks has evolved as their accuracy on ImageNet has continued to improve. We measure adversarial robustness in two different ways: First, we measure the smallest adversarial attack needed to cause a model to change its object categorization decision. Second, we measure how aligned successful attacks are with the features that humans find diagnostic for object recognition. We find that adversarial attacks are inducing bigger and more easily detectable changes to image pixels as DNNs grow better on ImageNet, but these attacks are also becoming less aligned with the features that humans find diagnostic for object recognition. To better understand the source of this trade-off and if it is a byproduct of DNN architectures or the routines used to train them, we turn to the \emph{neural harmonizer}, a DNN training routine that encourages models to leverage the same features humans do to solve tasks~\cite{Fel2022-dt}. Harmonized DNNs achieve the best of both worlds and experience attacks that are both detectable and affect object features that humans find diagnostic for recognition, meaning that attacks on these models are more likely to be rendered ineffective by inducing similar effects on human perception. Our findings suggest that the sensitivity of DNNs to adversarial attacks can be mitigated by DNN scale, data scale, and training routines that align models with biological intelligence. We release our code and data to support this goal.
\end{abstract}

\section{Introduction}
\footnotetext[1]{These authors contributed equally.}
\footnotetext{$^{1}$Department of Cognitive, Linguistic, \& Psychological Sciences, Brown University, Providence, RI}
\footnotetext{$^{2}$Carney Institute for Brain Science, Brown University, Providence, RI}
\footnotetext{$^{3}$Department of Computer Science, Brown University, Providence, RI}
\footnotetext{$^{4}$Institut de Recherche Technologique Saint-Exup\'ery, Toulouse, France}
\footnotetext{$^{5}$Artificial and Natural Intelligence Toulouse Institute, Toulouse, France}

For at least a decade, it has been known that the behavior of deep neural networks (DNNs) can be controlled by small ``adversarial'' perturbations of the input that are imperceptible to humans~\cite{Szegedy2014-qj,Biggio2017-ic}. As DNNs are increasingly being incorporated into software and tools we use in our everyday lives, their vulnerability to adversarial attacks is potentially an unsolved existential threat to the security and safety of these architectures. However, over recent years, the danger of adversarial attacks has been overshadowed by the ever-increasing scale-up of DNNs, and their resulting remarkable achievements across vision, language, and robotics. Billion-parameter DNNs are being trained on internet-scale datasets to perform tasks at levels that rival or surpass humans, bringing us tantalizingly closer to intelligent systems that can transform our lives for the better. It is not known how the scale of DNNs has affected their sensitivity to adversarial attacks.

\begin{figure}[h!]
\begin{center}
   \includegraphics[width=.99\linewidth]{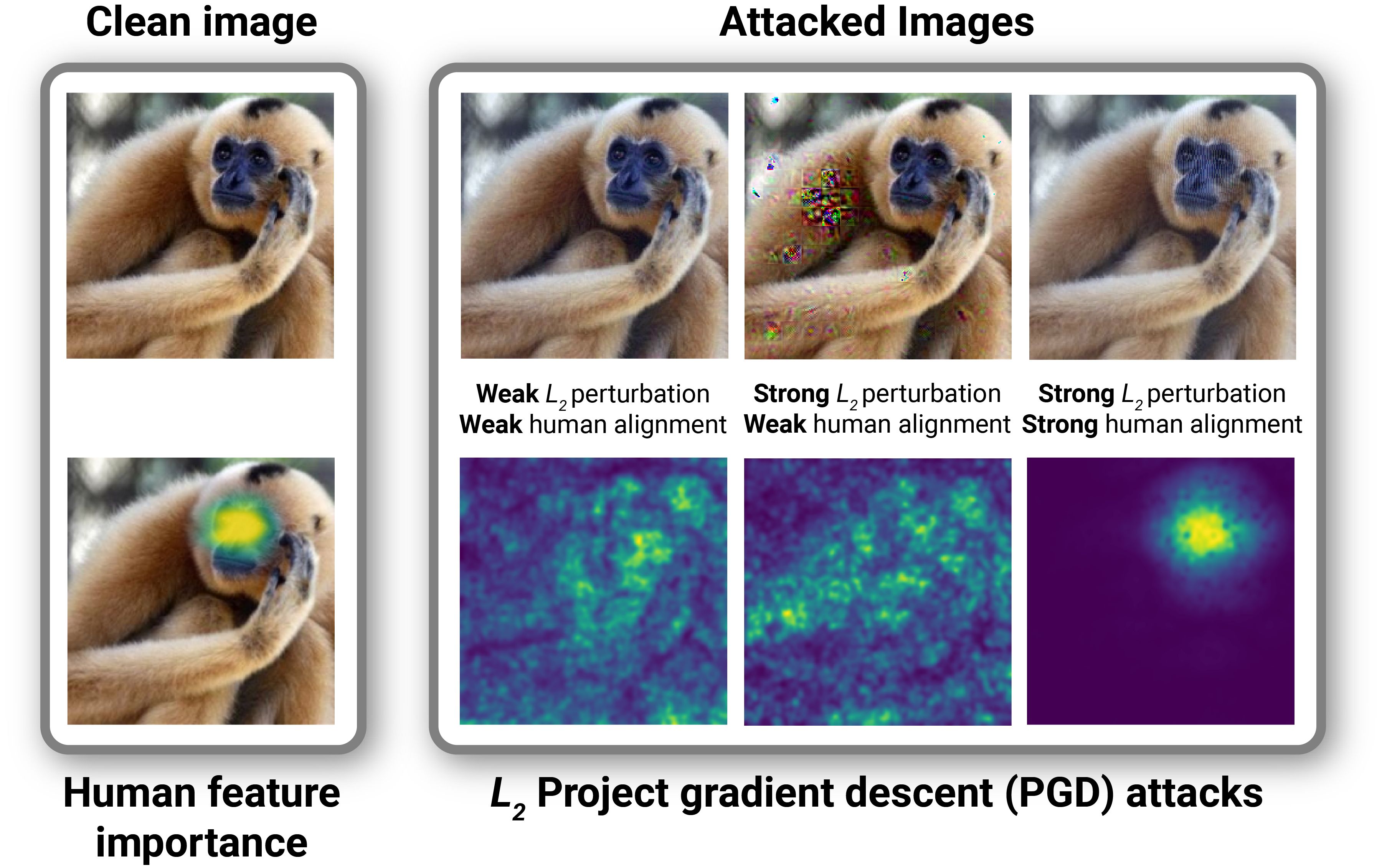}
\end{center}
   \caption{\textbf{We propose a new goal for adversarial robustness: Robust models are not only tolerant to strong adversarial image perturbations, successful attacks also target object features that humans find diagnostic for classification.} Adversarial attacks that are large in size and aligned with human perception are more likely to affect humans like they do models, which will neutralize their effectiveness. Shown here are an image of a snow monkey, its corresponding human feature importance map from \emph{ClickMe}~\cite{Linsley2019-xi}, and ``untargeted'' adversarial attacks from $\ell_2$ projected gradient descent (PGD) on three different DNNs. One DNN can be attacked with a weak perturbation (as measured by $\ell_2$ distance between clean and attacked images), and the successful attack is misaligned with the \emph{ClickMe} feature importance map according to the Spearman correlation between the two. Another DNN is more tolerant to perturbations, but successful perturbations are still misaligned with human perception (strong perturbation/weak alignment). A third DNN approaches our ideal result: strong perturbations are needed for successful attacks, and those perturbations affect features humans use for recognizing the object (the face), which renders these attacks more easily detectable and less effective. Zoom in to see details of each attack.}
\label{fig:teaser}
\end{figure}

There are a number of known ways to make DNNs more ``robust'' to adversarial attacks, meaning that it will take a larger change in pixels between an attacked and clean image to trick a model~\cite{kurakin2018adversarial}. For example, there are algorithmic defenses that can be incorporated into DNN inference~\cite{cohen2019certified} and training routines that increase the adversarial robustness of DNNs~\cite{PGD_madrytowards, trades_zhang2019theoretically,cisse2017parseval}. These approaches carry two key drawbacks. First, there is a well-established trade-off between a model's adversarial robustness and its task accuracy~\cite{yang2020closer,stutz2019disentangling}. Second, while improving a DNN's robustness means that a stronger perturbation is needed to attack it, there is no constraint on what parts of images are attacked. Humans rely on certain features more than others to recognize objects~\cite{Fel2022-dt,Linsley2019-xi,Schyns1994-ox,Ullman2016-wy}, and if a DNN attack affects features that are less important to humans for recognition, it may still prove be ignored or difficult to notice~\cite{Malhotra2020-kj} regardless of the perturbation strength (Fig.~\ref{fig:teaser}). We propose that for DNNs to be truly robust to adversarial attacks, then perturbations should induce large and detectable changes to the object features that humans find diagnostic for recognition (Fig.~\ref{fig:teaser}).

There is reason to believe that the scaling laws which have helped DNNs reach their many recent successes in vision and language may at least partially improve their adversarial robustness~\cite{bubeck2023universal}. Large-scale vision transformers are as robust to non-adversarial image perturbations as humans are, and it is possible that this means larger adversarial perturbations are needed to attack these models~\cite{Dehghani2023-zw,Geirhos2021-rr}. However, DNNs with high accuracy on ImageNet are also learning to recognize objects with features that are misaligned with those used by humans~\cite{Fel2022-dt}. It is not clear how these features of large-scale DNN vision interact and whether or not they affect the adversarial robustness of models.

\paragraph{Contributions.} 
In this work, we evaluate a large and representative sample of DNNs from the past decade to understand how their adversarial robustness has changed as they have evolved and improved on ImageNet.  We measure adversarial robustness in two ways: (\textit{i}) the average $\ell_{2}$ distance between clean and attacked images, which we refer to as ``perturbation tolerance'', and (\textit{ii}) the alignment of attacks with object features that humans find diagnostic for recognition, which we refer to as ``adversarial alignment''. We discover the following:
\begin{itemize}[leftmargin=*]\vspace{-2mm}
    \item DNNs have experienced a significant increase in perturbation tolerance as they have improved on ImageNet. In other words, the scale-up of DNNs that has happened over the past several years has partially helped defend them from adversarial attacks.
    \item In contrast, successful attacks on DNNs are becoming significantly less aligned with the object features that humans rely on for recognition~\cite{Linsley2020-en,Fel2022-dt} as models grow more accurate at ImageNet.
    \item Vision transformers~\cite{Dehghani2023-zw,Dosovitskiy2021-zy} (ViTs) and convolutional neural networks (CNNs) are robust in significantly different ways: ViTs have greater perturbation tolerance but CNNs have better adversarial alignment. Most importantly, there is a pareto-front governing the trade-off between these ways of measuring adversarial robustness, indicating that new approaches are needed for human-like adversarial robustness.
    \item We achieve a partial solution to this goal with the \emph{neural harmonizer}~\cite{Fel2022-dt}, a routine for aligning DNN representations with humans that significantly improves perturbation tolerance and adversarial alignment. 
\end{itemize}

\section{Methods}
\paragraph{DNN model zoo.}
We measured the adversarial robustness of 283 DNNs, which are representative of the variety of approaches used in computer vision today. There were 127 {\color{CNN}{convolutional neural networks}} (CNNs) trained on ImageNet~\cite{Chen2021-is,Tan2019-uh,Radosavovic2020-cs,Howard2019-cr,Simonyan2014-jd,Huang2018-yt,He2015-lm,Zhang2020-my,Gao2021-er,Kolesnikov2019-gg,Sandler2018-lh,Liu2022-es,Szegedy2016-fd,Szegedy2015-pr,Chollet2016-np,Radford2021-km,Xie2019-ju,Xie2016-ol,Szegedy2015-pr,Brendel2019-mw,Mehta2020-ad,Chen2017-wp,Wang2019-jm,Tan2018-zk, Xie2019-rp,Radford2021-km,Liu2022-es}, 123 {\color{transformer}{vision transformers}}~\cite{Dehghani2023-zw,Dosovitskiy2021-zy,Touvron2020-fo,El-Nouby2021-iw,Liu2021-cv,Liu2021-sj,Mehta2021-dd,Tu2022-br,Fang2023-yq,Fang2022-sx,Touvron2021-zu} (ViT), and 15 {\color{hybrid}{CNN/ViT hybrid architectures}} that used a combination of both types of circuits~\cite{Xu2021-oy,Yuan2021-vh}. Each model was implemented in PyTorch with the TIMM toolbox (\url{https://github.com/huggingface/pytorch-image-models}), using pre-trained weights downloaded from TIMM. Additional details on these DNNs, including the licenses of each, can be found in Appendix~\textsection{A}.

\paragraph{Neural Harmonizer.} There is a growing body of work indicating that the representations and perceptual behaviors of DNNs are becoming less aligned with humans as they improve on ImageNet~\cite{Fel2022-dt,Kumar2022-lv,Bowers2022-cy}. It has also been found that this misalignment can be partially addressed by the \emph{neural harmonizer}, a training routine that forces DNNs to learn object recognition using features that are diagnostic for humans. As this approach has significantly improved the alignment of DNNs with humans~\cite{Fel2022-dt}, we hypothesized that it would also improve the adversarial alignment of DNNs without inhibiting their ability to accurately recognize objects.

Training DNNs for ImageNet with the \emph{neural harmonizer} involves adding an another loss to cross-entropy for object recognition optimization. The additional loss forces a model's gradients to be as similar as possible to feature importance maps collected from humans. Distances between DNN and human feature imporance maps are computed at multiple scales by a function $\pyramid_i(.)$, which downsamples each map $\vp$ to $N$ levels of a pyramid using a Gaussian kernel, with $i \in \{1, ..., N\}$. To train a DNN with the \emph{neural harmonizer} we seek to minimize $\sum_{i}^N || \pyramid_i(\explainer(\pred, \vx)) - \pyramid_i(\gt) ||^2$ and align DNN feature importance maps with humans at every level of the pyramid. To facilitate learning, feature importance maps from DNNs and humans are normalized and rectified before distances are computed using $\std(.)$, a preprocessing function that takes a feature importance map $\gt$ and transforms it to have 0 mean and unit standard deviation. Putting these pieces together, the completed \emph{neural harmonizer} loss involves computing the following:

\begin{align}
    \mathcal{L}_{\text{Harmonization}} =&   
    \lambda_1 \sum_{i}^N || (\std \circ \pyramid_i \circ \explainer(\pred, \vx))^+ - (\std \circ \pyramid_i(\gt))^+  ||_2 
    \\& + \mathcal{L}_{CCE}(\pred, \vx, \vy) 
    + \lambda_2 \sum_{i} \theta_i^2
\label{eq:meta}
\end{align}

We follow the original \emph{neural harmonizer} training recipe to optimize 14 DNNs for object recognition on ImageNet while relying on category-diagnostic features captured by \emph{ClickMe}~\cite{Fel2022-dt}: one \texttt{VGG16}, one \texttt{ResNet50$\_$v2}, one \texttt{ViT$\_$b16}, one \texttt{EfficientNet$\_$b0}, six versions of \texttt{ConvNext Tiny}, and four versions of \texttt{MaxViT Tiny}. Each version was trained with different settings of $\lambda_1$ and $\lambda_2$, which controlled the relative strength of losses for object recognition and alignment, respectively (see Appendix~\textsection{B} for details).

\begin{figure}[h!]
\begin{center}
   \includegraphics[width=.99\linewidth]{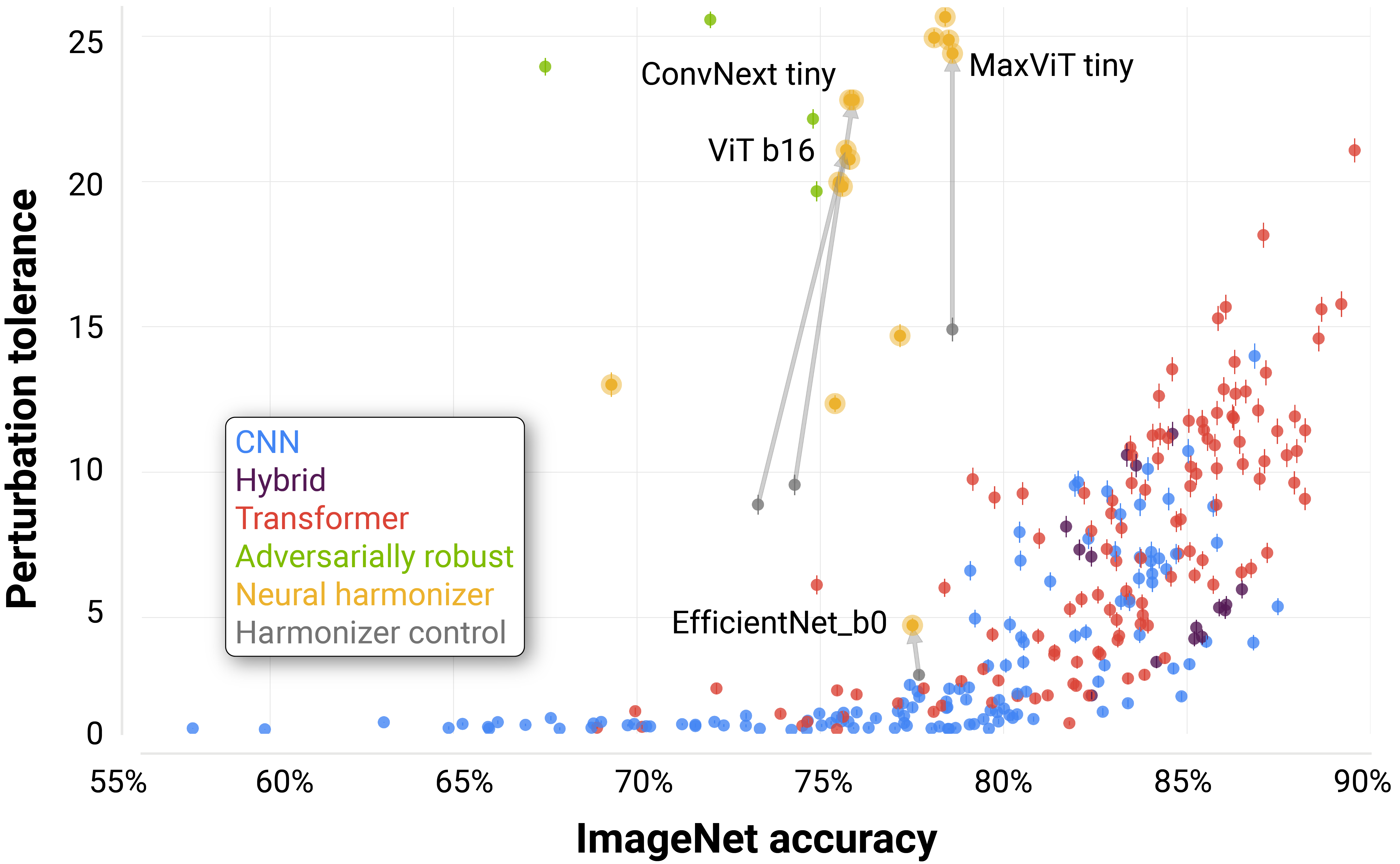}
\end{center}
   \caption{\textbf{The perturbation tolerance of DNNs has significantly increased as they have improved on ImageNet.} Each dot denotes a DNN's ImageNet accuracy vs. its average $\ell_2$ robustness radius to $\ell_2$ PGD attacks, which we call ``perturbation tolerance''. Arrows show the change of a DNN in both dimensions after it has been trained with the neural harmonizer. There is a significant positive correlation between ImageNet accuracy and perturbation tolerance ($\rho_{s} = 0.70$, $p < 0.001$). Error bars denote standard error, and variance may be so small for some models that they are not visible.}
\label{fig:l2}
\end{figure}

\emph{ClickMe} is a large-scale effort for capturing feature importance maps from human participants that highlight parts of objects that are relevant and irrelevant for recognition. For example, these maps focus on the faces of animals, the wheels and fronts of cars, and the wings and cockpits of airplanes~\cite{Linsley2019-ew}. Models were trained for object recognition using the ImageNet training set and \textit{ClickMe} human feature importance maps for the nearly 200,000 images that had annotations. The training was done on Tensorflow 2.0 with 8 V4 TPU cores per model. An object recognition loss was computed for every image, and the full harmonization loss was only computed for those images that had human feature importance maps. Batches of 512 images and feature importance maps were augmented with random left-right flips and mixup~\cite{Zhang2017-hw} during training. Model weights were optimized using SGD with momentum, label smoothing~\cite{Muller2019-td}, a learning rate of $0.3$, and a cosine learning rate schedule consisting of a five epoch warm-up followed by learning rate decays at steps $30$, $50$, and $80$. We also trained versions \texttt{ViT$\_$b16}, \texttt{EfficientNet$\_$b0}, \texttt{ConvNext Tiny}, and \texttt{MaxViT Tiny} with crossentropy but not the complete \emph{neural harmonizer} as controls, and refer to these as {\color{baseline}{harmonizer control}} models.

\paragraph{Adversarially robust DNNs}
We also tested the perturbation tolerance and adversarial alignment of robust DNNs. We trained four \texttt{Robust ResNetv2-50s} to be tolerant to $\ell_{\infty}$-bounded attacks using a standard procedure~\cite{PGD_madrytowards,Salman2020-lo} (code from \url{https://github.com/microsoft/robust-models-transfer}). A DNN's robustness to these attacks is controlled by a hyperparameter $\epsilon$, and we trained versions with $\epsilon \in {0.001, 0.01, 0.05, 0.1}$ and the same training setup used for models trained with the \emph{neural harmonizer}.

\paragraph{Experimental stimuli.}
We selected 1000 images at random from the ImageNet validation set which also had \emph{ClickMe} feature importance maps. Each image was from a different ImageNet category, and images were preprocessed with each model's specific procedure before computing adversarial attacks.

\paragraph{Adversarial attacks.} Ever since the introduction of adversarial attacks~\cite{Szegedy2014-qj}, the field has exploded with variations that trade-off speed for effectiveness. In our study, we were interested in using attacks that (\textit{i}) could be applied to our model zoo and stimulus set in a reasonable amount of time, (\textit{ii}) would approach the smallest perturbation needed to change a model's behavior, and (\textit{iii}) yielded continuous-valued perturbations that could be compared to \emph{ClickMe} feature importance maps to measure their alignment with human perception. One candidate for these criteria is the popular Fast Gradient Sign Method (FGSM~\cite{FGSM_goodfellow2014explaining}), which is renowned for its time efficiency. However, its attacks are suboptimal~\cite{PGD_madrytowards}, and it belongs to the $L_\infty$ attack category which only captures the sign of an attack at every pixel and is poorly suited for computing correlations with human feature importance maps. We instead turned to $\ell_2$ Projected Gradient Descent (PGD~\cite{PGD_madrytowards}), which fits each of our criteria. $\ell_2$ PGD iteratively searches for the smallest possible image perturbation within a fixed $\epsilon$ ball that changes model behavior.

For each model in our zoo, we ran a single $\ell_2$ PGD attack for 3 iterations and used binary search to find the minimum perturbation tolerance with $\epsilon \in {0.001, 10}$ on every ImageNet image in our stimulus dataset. We also constrained attacks to fall within the pixel range of natural images (i.e. $[0, 255]$). We report perturbation tolerance as the $\ell_2$ distance between a clean version of an image and the minimum $\epsilon$ attacked-version. All attacks were successful for every image and model. We used NVIDIA TITAN 8 GPUs for generating adversarial attacks, which took between 30 and 240 minutes per model for the complete 1000-image stimulus dataset.

\begin{figure}[h!]
\begin{center}
   \includegraphics[width=.99\linewidth]{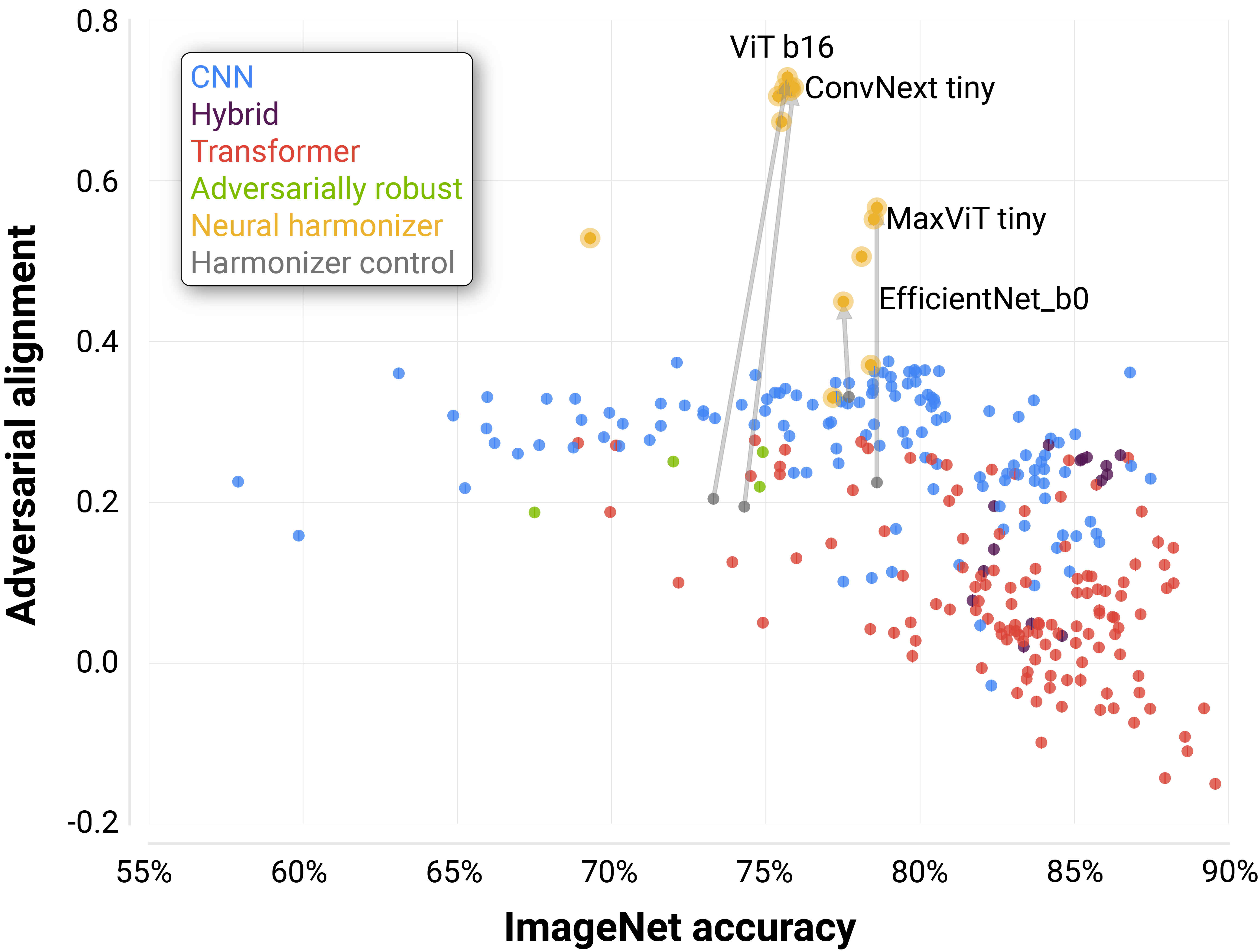}
\end{center}
   \caption{\textbf{Successful adversarial attacks on DNNs are becoming less aligned with human perception as they have improved on ImageNet.} Each dot denotes a DNN's ImageNet accuracy vs. the average Spearman correlation between successful attacks an images' human feature importance maps from \emph{ClickMe}. We call this correlation a DNN's adversarial alignment. Arrows show the change of a DNN in both dimensions after it has been trained with the neural harmonizer. Error bars denote standard error, and variance may be so small for some models that they are not visible.}
\label{fig:alignment}
\end{figure}

\section{Results}
\paragraph{DNNs are becoming more tolerant to adversarial attacks as they improve on ImageNet.} We used $\ell_2$ PGD to attack the object recognition decisions of every DNN in our model zoo for the 1000 images in our stimulus set. We computed perturbation tolerance scores for each DNN as the average $\ell_2$ distance between clean images and the attacked versions found by PGD that changed its recognition decision. Surprisingly, as DNNs have improved on ImageNet, their perturbation tolerance has also improved, significantly (Fig.~\ref{fig:l2}, $\rho_{s} = 0.70$, $p < 0.001$). As a point of comparison, the most accurate DNN we tested, the \texttt{eva\_giant\_patch14\_336.m30m\_ft\_in22k\_in1k}, rivaled the perturbation tolerance of \texttt{Robust ResNetv2-50s} (i.e., trained for perturbation tolerance) despite being approximately 22\% more accurate on ImageNet. We also found a shift in perturbation tolerance based on model architecture. ViTs were significantly more tolerant to perturbations than CNNs (Fig.~\ref{fig:l2}, red vs. blue, $T(122) = 9.12$, $p < 0.001$). We found that this pattern of results replicated when using $\ell_\infty$ PGD instead of $\ell_2$ PGD (Appendix~\textsection{C}, $\rho_{s}=0.72$, $p < 0.001$). In other words, the continued optimization of DNNs for performance on ImageNet holds promise for building models that are as robust to image perturbations as any approach designed specifically to build such tolerance.

\begin{figure}[h!]
\begin{center}
   \includegraphics[width=.99\linewidth]{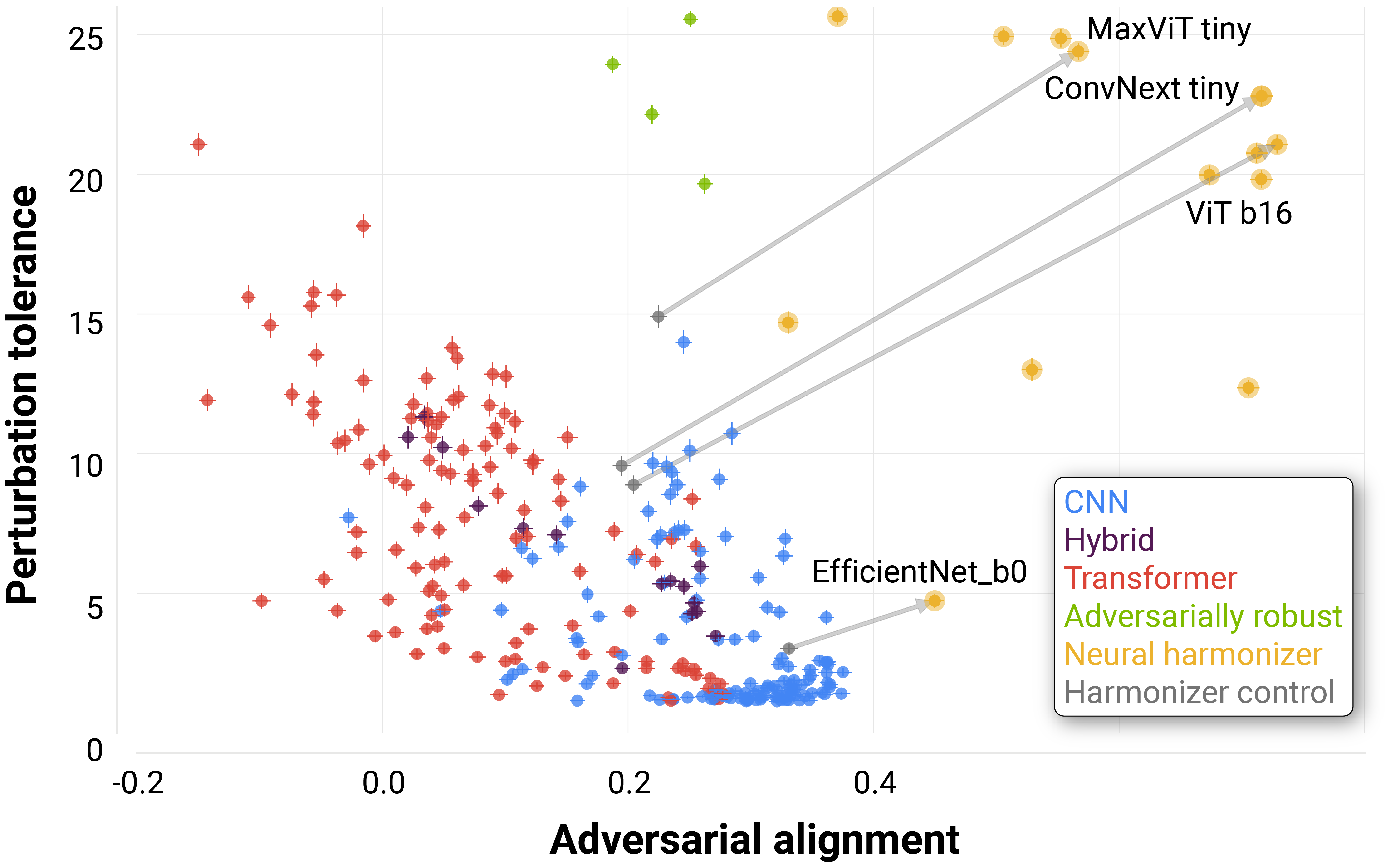}
\end{center}
   \caption{\textbf{DNNs trade-off between adversarial alignment perturbation tolerance.} Each dot denotes a DNN's average Spearman correlation between successful attacks and images' human feature importance maps from \emph{ClickMe} vs. the $\ell_2$ distance between successfully attacked and clean images. We call these scores adversarial alignment and perturbation tolerance, respectively. Arrows show the change of a DNN in both dimensions after it has been trained with the neural harmonizer. Error bars denote standard error, and variance may be so small for some models that they are not visible.}
\label{fig:tradeoff}
\end{figure}

\paragraph{Successful adversarial attacks are becoming less aligned with human perception.} We propose that an adversarially robust DNN should not only be tolerant to strong image perturbations, successful attacks should also target features that humans find diagnostic for object recognition. In this way, even if an attack is successful, it will affect humans like it does DNNs, making the image more difficult to recognize and reducing the potency of the attack. To measure the alignment of a model's adversarial attacks with humans, we turned to \emph{ClickMe}, a large-scale dataset of human feature importance maps for ImageNet~\cite{Linsley2019-ew}. We then measured a DNN's adversarial alignment with humans as the average Spearman correlation between \emph{ClickMe} maps and successful adversarial attacks for every image in our stimulus set.

As DNNs have improved on ImageNet, the alignment of their attacks with human perception has dropped significantly (Fig.~\ref{fig:alignment}, $\rho_{s} = -0.53, p < 0.001$). The 89.57\% accurate \texttt{eva\_giant\_patch14\_336.m30m\_ft\_in22k\_in1k} has a $\rho_{s} = -0.15$ adversarial alignment with humans, whereas the 78.98\% accurate \texttt{MixNet-L} has a $\rho_{s} = 0.38$ adversarial alignment with humans. In contrast to our findings with perturbation tolerance, CNNs were on average significantly more adversarially aligned with humans than ViTs (Fig.~\ref{fig:alignment}, red vs. blue, $T(122) = -18.73$, $p < 0.001$).

\paragraph{DNNs trade-off between perturbation tolerance and adversarial alignment.} After plotting the perturbation tolerance of each DNN in our zoo against its adversarial alignment, we found a striking pattern: DNNs either have a strong tolerance to perturbations and misaligned attacks \emph{or} successful attacks are weak in strength but moderately aligned with human perception. The partial outlier to this pattern is DNNs trained for adversarial robustness, which are tolerant to strong perturbations and have moderate adversarial alignment, but are also relatively inaccurate on ImageNet.

\begin{figure}[h!]
\begin{center}
   \includegraphics[width=.99\linewidth]{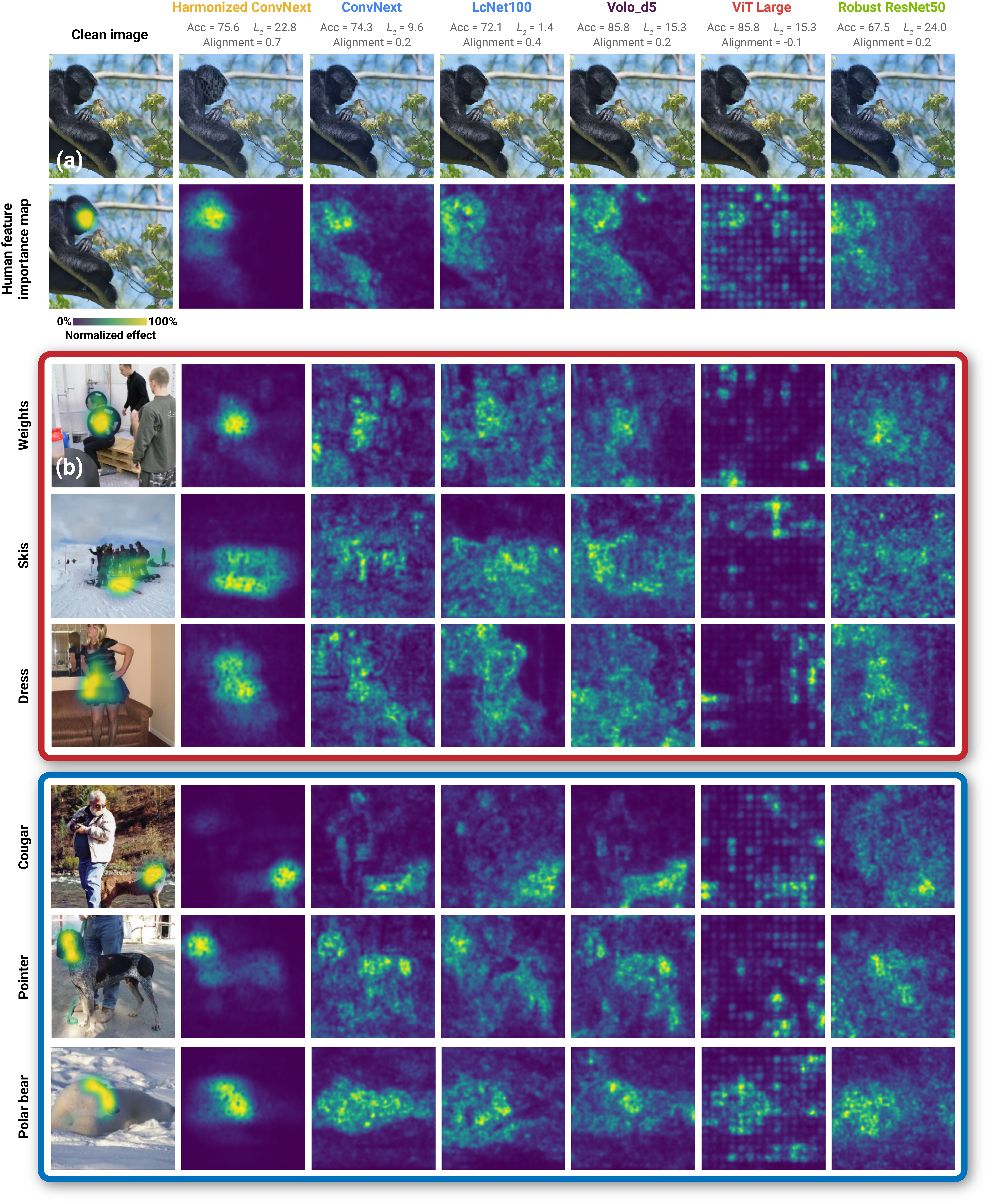}
\end{center}
   \caption{\textbf{$\ell_2$ PGD adversarial attacks for DNNs.} Plotted here are ImageNet images, human feature importance maps from \emph{ClickMe}, and adversarial attacks for a variety of DNNs. Attacked images are included for the image of a monkey at the top (zoom in to see attack details). The red box shows inanimate categories, and the blue box shows animate categories.}
\label{fig:qualitative}
\end{figure}

We reasoned that another approach for breaking the perturbation strength and adversarial alignment trade-off we observed is to train models for alignment with human perception. One solution to this problem is the \emph{neural harmonizer}, which can significantly improve the representational alignment of DNNs with human perception while also maintaining or slightly improving model accuracy on ImageNet~\cite{Fel2022-dt}, unlike adversarial robustness training~\cite{PGD_madrytowards}. Indeed, we found that a \texttt{harmonized ResNet50} approached an \texttt{adversarially-trained ResNet50} in average perturbation tolerance (14.69 vs. 23.95) while being 14\% more accurate on ImageNet (Fig.~\ref{fig:l2}). We also observed that successful attacks on harmonized DNNs were significantly more aligned with human perception than any other DNN, and they break the perturbation tolerance and adversarial alignment trade-off faced by nearly all other DNNs (average alignment of harmonized DNNs vs. the most aligned unharmonized DNN, $T(999) = 15.63$, $p < 0.001$, Fig.~\ref{fig:tradeoff}). 

Successful adversarial attacks on harmonized models target features that humans rely on for recognition: for example, distorting the face of a monkey but leaving the rest of an image untouched (Fig.~\ref{fig:qualitative}). All other DNNs, including ones trained for adversarial robustness, have attacks that affect image context as much or more than they do the foreground object. While large-scale and highly accurate DNNs like the ViT have high perturbation tolerance, meaning that successful attacks can be visible and detectable by eye (Appendix~Fig.~S1-2), these patterns of noise may be ignored as inconsequential image distortions~\cite{Dujmovic2020-rv} since they rarely affect features that are diagnostic for humans. 

\section{Related work}
\paragraph{Adversarial attacks and human perception.} Adversarial attacks represent a major threat to safety and security because they are hard or impossible to detect by eye. This feature of adversarial attacks -- their perceptibility or lack thereof -- has also made them a popular source for study in the vision sciences. It has been suggested that even though adversarial attacks look nonsensical, humans can nevertheless decipher their meaning~\cite{Zhou2019-lf}. Similarly, there is evidence that adversarial attacks on CNNs can transfer to humans in rapid psychophysics experiments~\cite{Elsayed2018-un} and that DNNs trained for adversarial robustness and neurons in primate inferotemporal cortex share a similar tolerance to adversarial perturbations~\cite{Guo2022-dp}. On the other hand, others have claimed that the similarities between the adversarial robustness of DNN and human vision can be arbitrarily controlled by experimental design and stimulus choices~\cite{Malhotra2020-kj,Dujmovic2020-rv,Malhotra2022-dl}. Our findings enrich and reconcile these disparate claims by demonstrating that adversarial robustness, as it is commonly used to describe perturbation tolerance, need not entail alignment with humans. DNNs that achieve perturbation tolerance and adversarial alignment will bring us one step closer towards artificial vision systems that see like humans do.

\paragraph{Aligning the visual strategies of humans and machines.} Taken to its extreme, it is possible that a DNN can have arbitrarily high tolerance to adversarial perturbations, but those perturbations could occur in a single, unnoticeable, pixel on the boundary of an image~\cite{Malhotra2020-kj}. This is one of the many reasons why there is a growing urgency in the field of computer vision to ensure that DNNs that rival human performance on image benchmarks can achieve their successes with visual strategies that are interpretable and at least partially consistent with those of humans. There has been progress made towards this goal by evaluating or co-training DNNs with data on human attention and saliency, gathered from eye tracking or mouse clicks during passive or active viewing~\cite{Linsley2017-qe,Linsley2019-ew,Jiang2015-vl,Lai2019-ln,Ebrahimpour2019-dc}. Others have achieved similar success by comparing the behaviors of models to humans, either by computing and optimizing for distances between patterns of behavior~\cite{Peterson2018-pu,Roads2020-gd,Sucholutsky2023-wm,Muttenthaler2022-hk}, or by combining behavioral data with human eye tracking~\cite{Langlois2021-ns}. Another direct comparison of human and DNN alignment involved identifying the minimal image patch needed by each for object recognition~\cite{Ullman2016-wy,Funke2018-ft}. While the \emph{ClickMe} data we used here for harmonizing DNNs is significantly larger than any of these other efforts, we suspect that they hold similar promise in helping DNNs improve the human perceptual alignment of adversarial attacks.

\section{Discussion}
\paragraph{DNN scale provides valuable protection against the strength of adversarial attacks.} Perhaps the biggest breakthrough in artificial intelligence since the release of AlexNet is the finding that scaling the number of parameters in DNNs and the size of their datasets for training can help them rival and outperform humans on challenging tasks~\cite{Kaplan2020-zx,Dehghani2023-zw}. Here, we show that scale also provides concomitant benefits to the perturbation tolerance of DNNs: the size of an adversarial attack needed to affect today's most largest-scale and most-accurate DNNs is significantly greater than ever before. This trend also appears to be accelerating, with ViTs growing tolerant at a faster rate than ever before. DNN scale may be sufficient for ``defanging'' adversarial attacks by making them detectable to humans.

\paragraph{DNN scale worsens their adversarial alignment with human perception.} As the perturbation tolerance of DNNs has improved with ImageNet accuracy, successful attacks on accurate models have begun to consistently affect parts of object images that humans find less important or completely irrelevant for recognition. In other words, DNN scale is at best only a partial solution to adversarial robustness, and it is important for the field of computer vision to explore new approaches to alignment to ensure that adversarial attacks target features humans rely on for behavior. Thus, even if adversarial attacks are successful, they will be ineffective because they induce the same behaviors in humans as they do in DNNs.

\paragraph{The \emph{neural harmonizer} is a short-term solution to adversarial robustness.} Harmonized DNNs achieve the best of both worlds of adversarial robustness: they have high perturbation tolerance, and successful attacks target features humans rely on for object recognition. We suspect that scaling the \emph{neural harmonizer} to larger and more accurate DNNs, and expanding the size of \emph{ClickMe} (potentially with pseudo-labels on internet-scale datasets), will bring the field closer to models that are sufficiently robust to adversarial attacks. The success of the \emph{neural harmonizer} also suggests that there is a fundamental misalignment of the training routines used for large-scale DNNs today, and it is possible that advances could also be made without \emph{ClickMe} feature importance maps by inducing more human-like developmental principles onto models. We release our code and data to support continued progress towards adversarial robustness (\url{https://serre-lab.github.io/Adversarial-Alignment/}).

\paragraph{Limitations.} We relied on $\ell_2$ PGD for our experiments because it is relatively fast and a ``universal first-order adversary''~\cite{PGD_madrytowards}, meaning that it is the strongest possible adversarial attack on a DNN that relies on first-order information. While this might suggest that our results are specific to $\ell_2$ PGD, we found they translate to $\ell_\infty$ PGD (Appendix~\textsection{C}). Moreover, in a very small-scale experiment we found a similar pattern of results with the highly-effective but extremely slow-to-compute Carlini-Wagner (CW) attack~\cite{CW_carlini2017towards} (Appendix~\textsection{C}). Thus, our findings are likely a general feature of adversarial attacks on DNNs.

\paragraph{Broader impacts.} Adversarial attacks have posed an immense problem for the security and safety of DNNs since their discovery. If a DNN's behavior can be controlled by an imperceptible pattern of noise added by a bad actor, then how can they ever be trusted in our everyday lives? We show that the scaling trends that are driving progress in computer vision today offer a partial solution to these attacks, and new approaches for inducing representational alignment between DNNs and humans can potentially close the remaining gap.

\begin{ack}
This work was supported by ONR (N00014-19-1-2029), NSF (IIS-1912280 and EAR-1925481), DARPA (D19AC00015), NIH/NINDS (R21 NS 112743), and the ANR-3IA Artificial and Natural Intelligence Toulouse Institute (ANR-19-PI3A-0004).
Additional support provided by the Carney Institute for Brain Science and the Center for Computation and Visualization (CCV). We acknowledge the Cloud TPU hardware resources that Google made available via the TensorFlow Research Cloud (TFRC) program as well as computing hardware supported by NIH Office of the Director grant S10OD025181.
\end{ack}

{\small
\bibliographystyle{splncs}
\bibliography{egbib, egbib_robustness}
}

\clearpage
\setcounter{figure}{0}
\makeatletter 
\renewcommand{\thefigure}{S\@arabic\c@figure}
\makeatother
\setcounter{table}{0}
\makeatletter 
\renewcommand{\thetable}{S\@arabic\c@table}
\renewcommand{\thefigure}{S\arabic{figure}}
\makeatother

\appendix

\section{DNN Model Zoo}\label{si_sec:dnn_model_zoo}
We comprehensively evaluated the adversarial robustness of DNNs on a large sample of models from the TIMM toolbox~\cite{rw2019timm}. These DNNs, available under the Apache 2.0 license, are intended for non-commercial research purposes. The complete list of DNNs we evaluated on can be in Table $~\ref{si_table:SI_model_zoo}$ below.

\begin{table}[h]
\centering
\begin{tabular}{ccc} 
\hline 
\textbf{Architecture}           & \textbf{Model} & \textbf{Versions}  \\ 
\hline
\multirow{18}{*}{CNN}   & VGG               & 8                \\
                        & ResNet            & 8                \\
                        & EfficientNet      & 7                \\
                        & ConvNext          & 6                \\
                        & MobileNet         & 10               \\
                        & Inception         & 3                \\
                        & DenseNet          & 4                \\
                        & RegNet            & 22               \\
                        & Xception          & 4                \\
                        & MixNet            & 4                \\
                        & DPN               & 6                \\
                        & DarkNet           & 1                \\
                        & NFNet             & 11               \\
                        & TinyNet           & 5                \\
                        & LCNet             & 3                \\
                        & DLA               & 12               \\
                        & MnasNet           & 4                \\
                        & CSPNet            & 3                \\ 
\hline
\multirow{8}{*}{ViT}    & General ViT       & 8                \\
                        & MobileViT         & 10               \\
                        & Swin              & 22               \\
                        & MaxViT            & 14               \\
                        & DeiT              & 24               \\
                        & CaiT              & 10               \\
                        & XCiT              & 28               \\
                        & EVA               & 5                \\ 
\hline
\multirow{2}{*}{Hybrid} & VOLO              & 8                \\
                        & CoAtNet           & 13               \\
\hline
\end{tabular}

\caption{ \label{si_table:SI_model_zoo} \textbf{A list of models selected from TIMM library.}}

\end{table}

%%%%%%%%%%%%%%%%%%%%%%%%%%%%%%%%%%%%%%%%%%%%%%%%%%%%%%%%%%%%

\section{\textit{Neural Harmonizer} Training}\label{si_sec:neural_harmonizer}

In our work, we followed the original neural harmonizer training recipe to train and harmonize 14 DNNs for object recognition on the ImageNet~\cite{Deng2009-jk} dataset. By adjusting the regularization terms $\lambda_1$ and $\lambda_2$, we controlled the relative importance of losses for object recognition and human feature alignment during training. We sampled as many $\lambda_1$ and $\lambda_2$ settings as possible given our resources, and included all versions in our experiments. In total, we trained one $\texttt{VGG16}$, one $\texttt{ResNet50\_v2}$, one $\texttt{ViT\_b16}$, one $\texttt{EfficientNet\_b0}$, six variations of $\texttt{ConvNext Tiny}$, and four variations of $\texttt{MaxViT Tiny}$ (Table ~\ref{si_table:neural_harmonizer}). Note that we did not attempt to harmonize models pre-trained on datasets other than ImageNet because the \emph{ClickMe} feature importance dataset we used contained annotations on a subset of images in ImageNet.

\begin{table}
\centering
\begin{tabular}{lccc} 
\hline
\textbf{Model}   & \textbf{Accuracy (\%)} & \textbf{Human Alignment (\%)} & \textbf{Note}  \\ 
\hline
VGG              & 69.3       & 61.5               & $\lambda$ = 2  \\
ResNet 50        & 77.17      & 45.0               & $\lambda$ = 2 \\
EfficientNet B0  & 77.51      & 52.3               & $\lambda$ = 20\\
ViT B16          & 75.7       & 72.6               & $\lambda$ = 5 \\
ConvNext Tiny v1 & 75.9       & 73.2               & $\lambda$ = 1 \\
ConvNext Tiny v2 & 75.8       & 73.3               & $\lambda$ = 2 \\
ConvNext Tiny v3 & 75.8       & 74.5               & $\lambda$ = 3 \\
ConvNext Tiny v4 & 75.5       & 72.1               & $\lambda$ = 5 \\
ConvNext Tiny v5 & 75.6       & 71.1               & $\lambda$ = 8 \\
ConvNext Tiny v6 & 75.4       & 73.2               & $\lambda$ = 10\\
MaxViT Tiny v1   & 78.6       & 45.3               & $\lambda$ = 1 \\
MaxViT Tiny v2   & 78.4       & 46.8               & $\lambda$ = 2 \\
MaxViT Tiny v3   & 78.5       & 57.6               & $\lambda$ = 5 \\
MaxViT Tiny v4   & 78.1       & 59.0               & $\lambda$ = 10 \\ 
\hline             
\end{tabular}
\caption{ \label{si_table:neural_harmonizer} \textbf{DNN architectures trained with the \emph{neural harmonizer.}}}
\end{table}

%%%%%%%%%%%%%%%%%%%%%%%%%%%%%%%%%%%%%%%%%%%%%%%%%%%%%%%%%%%%%%%%%%%%

\section{Adversarial attacks}\label{si_sec:adv_attacks}
\paragraph{$\ell_2$ PGD.} The core idea of the Projected Gradient Descent (PGD) attack is to cast adversarial attacks as a constrained optimization problem. PGD leverages the (first-order) gradient information of the model to optimize adversarial attacks while keeping the perturbation size $\delta$ within certain constraints. Each step of the PGD attack that we used in our experiments can be presented as follows:

\begin{equation}
    \delta:=\mathcal{P}\left(\delta+\alpha \nabla_{\delta} \operatorname{loss}\left(f_{\theta}(x+\delta), y\right)\right)
\end{equation}

, where $f_{\theta}$ refers to DNN, $\alpha$ means step size, and $\mathcal{P}$ denotes the projection onto the ball of interest. In our experiments, we used the $\ell_2$ PGD to attack the object recognition decisions of DNNs. This attack aims to generate adversarial images by perturbing the input data within a bounded $\ell_2$ norm or Euclidean ball, iteratively moving an attacked image representation to the closest point on the circle of a particular radius $\epsilon$ centered at the image representation origin ( i.e $ \| \delta \|_2 = \epsilon $).

To obtain the perturbation tolerance of DNNs, the objective is to find the minimum $\epsilon$ that causes the failure of model identification for each image. In our approach, we iteratively refined the value of $\epsilon$ using binary search, to efficiently find the minimum $\epsilon$. We started by setting a lower-bound epsilon value ${\epsilon}_{l}$ and an upper-bound epsilon value ${\epsilon}_{u}$ based on empirical knowledge. The lower-bound value represents the minimum perturbation that we assumed could result in misclassification, while the upper-bound value was initially set to a large value that we expected would always result in model failure. In each iteration of the binary search, we perturbed the clean image with a midpoint value between ${\epsilon}_{l}$ and ${\epsilon}_{u}$. Then, we evaluated the generated adversarial example by feeding it into the DNN model. If the model correctly identified the adversarial examples, we adjusted the lower-bound epsilon value ${\epsilon}_{l}$ to the midpoint. However, if the model made the wrong prediction, we updated the upper-bound epsilon value ${\epsilon}_{u}$ to the midpoint, narrowing down the search range accordingly. We repeated this process until the difference between the upper-bound and lower-bound epsilon values was less than a predefined threshold, indicating that we had converged to a minimum perturbation tolerance of that single image. At this point, we averaged the $\ell_2$ distortions between the clean image and their corresponding adversarial example. The pseudocode is shown in Alg.~\ref{alg:mpt}.  

\begin{algorithm}[t]
\textbf{Input:} {
\hspace*{0.1em} a DNN $\mathcal{F}_{\theta}$, image-label pairs $(\mathcal{X}, \mathcal{Y})$, $\ell_2$ PGD function $\mathcal{P}$, $\ell_2$ norm function $\mathcal{N}$, \newline
\hspace*{3.5em} lower-bound epsilon ${\epsilon}_{l}$, upper-bound epsilon ${\epsilon}_{u}$, threshold $k$.}

\textbf{Output:} {minimum perturbation tolerance $t$.}

\begin{algorithmic}
% \State $\mathcal{T} \gets \text{empty list}$
\State $\mathcal{T} \gets \text{[\ ]}$
\For{$x_i, y_i$ in $\mathcal{X}, \mathcal{Y}$} 
    \State $l, r \gets {\epsilon}_{l}, {\epsilon}_{u}$

    \While{$r - l \geq k$}
        \State $m \gets l + (r - l) / 2$ 
        \State $\hat{x}_i, \hat{y}_i \gets \mathcal{P}(x_i, y_i, m, \mathcal{F}_{\theta})$
        
        \If{$y_i \neq \hat{y}_i$}
            \State $r \gets m$ 
        \Else 
            \State $l \gets m$ 
        \EndIf
    \EndWhile
    
    \State $\epsilon \gets r$
    \State $\hat{x}_i, \hat{y}_i \gets \mathcal{P}(x_i, y_i, m, \mathcal{F}_{\theta})$
    \State $\mathcal{T}.\operatorname{append}(\mathcal{N}(x_i, \hat{x}_i))$
\EndFor

\State $t \gets \operatorname{mean}(\mathcal{T})$

\State \Return {t}

\end{algorithmic}
\caption{\label{alg:mpt}Find the minimum perturbation tolerance of an DNN model.}
\end{algorithm}

%%%%%%%%%%%%%%%%%%%%%%%%%%%%%%%%%%%%%%%%%%%%%%%%%%%%%%%%%%%%

\paragraph{$\ell_2$ PGD Attack on Large-scale DNNs.}\label{si_sec:l2_large_dnn}

Large-scale and highly accurate DNNs, especially ViT-based models, exhibit a high perturbation tolerance; successful attacks can often result in visible perturbations. However, these visually noticeable patterns of noise do not always affect image features that humans rely on for recognizing objects.

\begin{figure}[t!]
\begin{center}
   \includegraphics[width=1\linewidth]{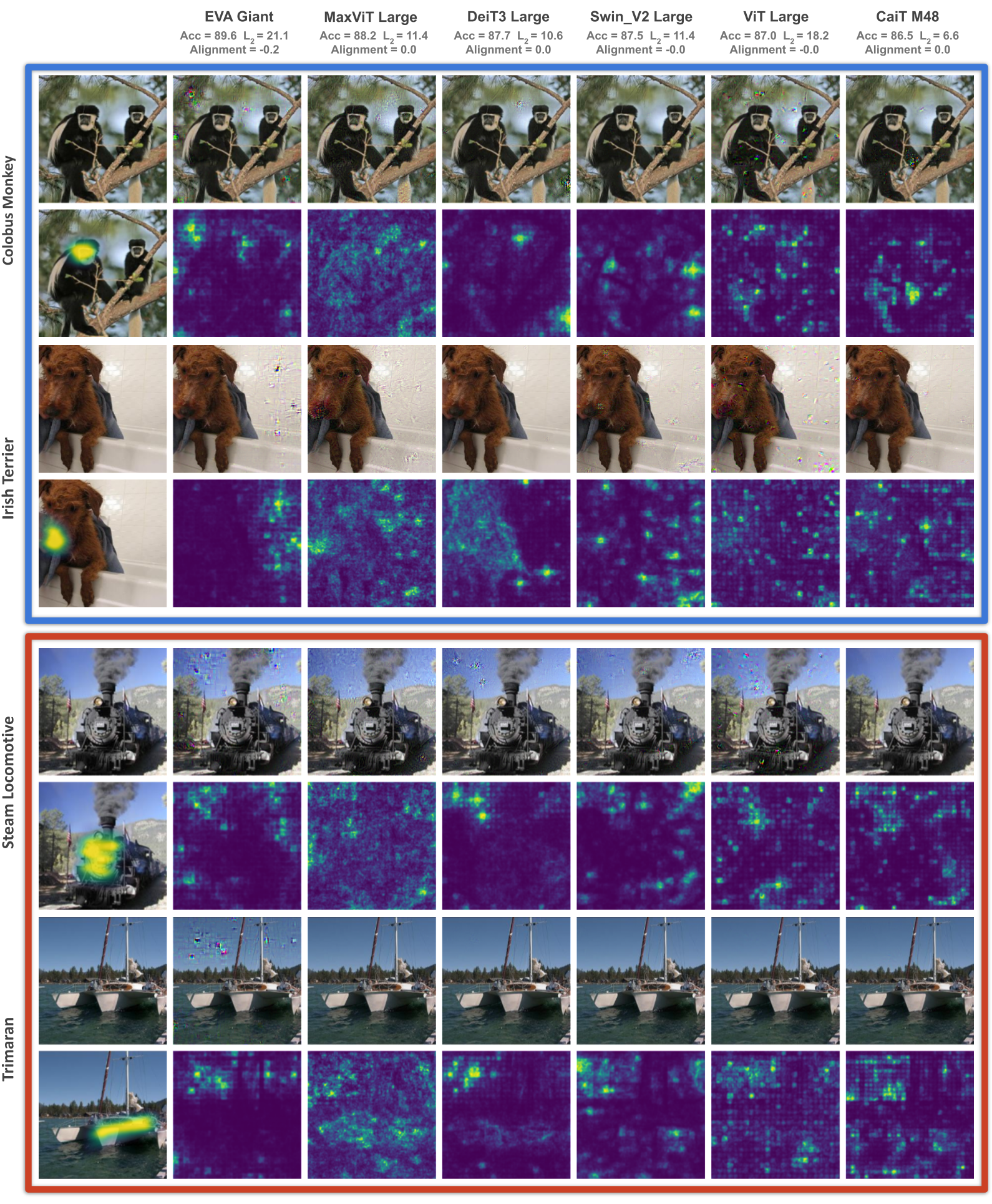}
\end{center}
   \caption{\textbf{$\ell_2$ PGD attack on large-scale models.} Plotted here are ImageNet images, human feature importance maps from ClickMe, and adversarial attacks for 6 large-scale and high-accuracy DNNs. The red box shows inanimate categories, and the blue box shows animate categories.}
\label{si_fig:SI_L2_large_DNN}
\end{figure}

\begin{figure}[t!]
\begin{center}
   \includegraphics[width=1\linewidth]{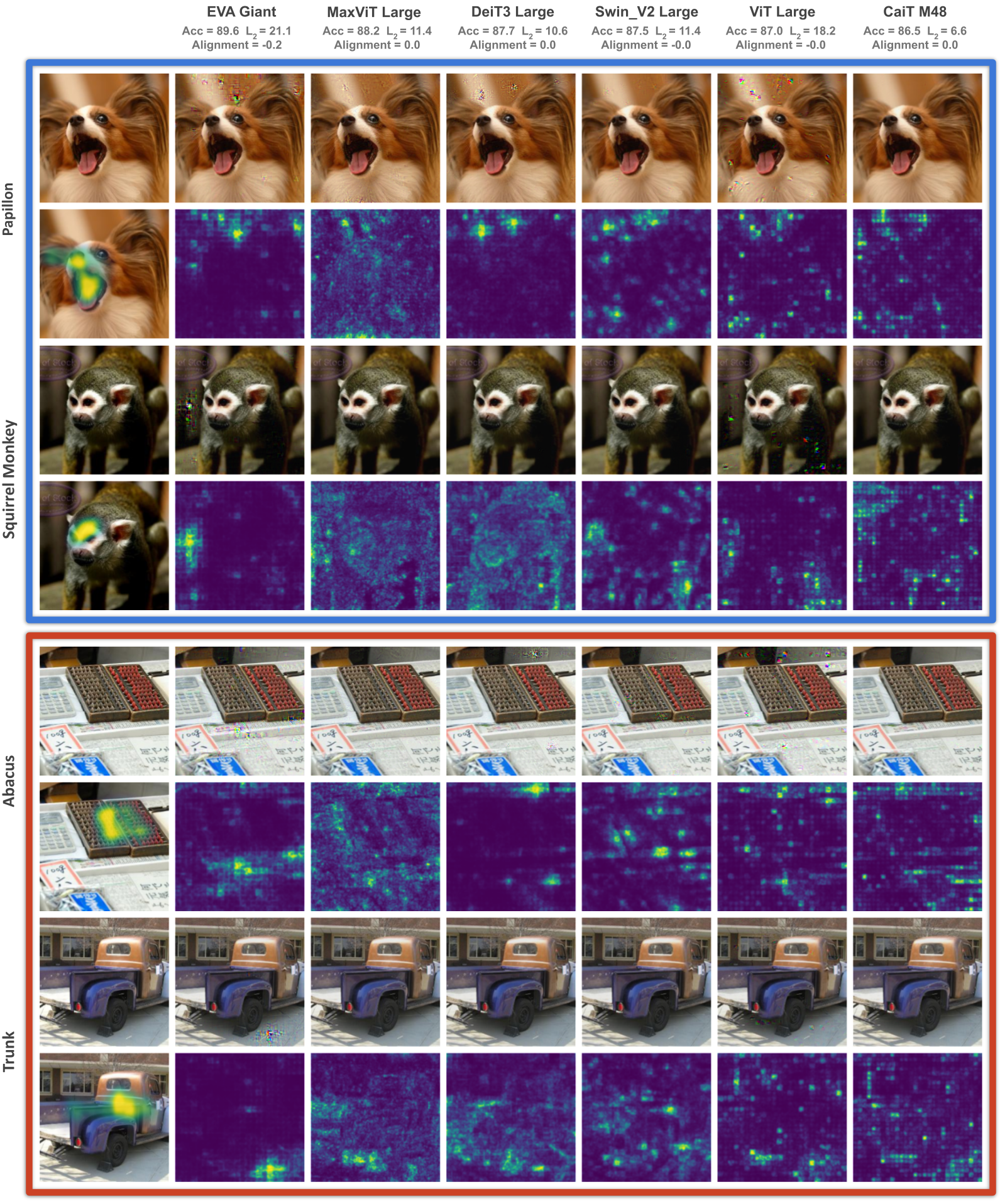}
\end{center}
   \caption{\textbf{$\ell_2$ PGD attack on large-scale models.} Plotted here are ImageNet images, human feature importance maps from ClickMe, and adversarial attacks for 6 large-scale and high-accuracy DNNs. The red box shows inanimate categories, and the blue box shows animate categories.}
\label{si_fig:SI_L2_large_DNN_2}
\end{figure}

Our study highlights the behavior of six large-scale ViT-based models that demonstrate exceptional performance on the ImageNet dataset, achieving Top-1 accuracy of over 86\% (Fig.$~\ref{si_fig:SI_L2_large_DNN}$ and Fig.$~\ref{si_fig:SI_L2_large_DNN_2}$). These DNNs exhibit high perturbation tolerance, but also have low alignment with human perception, complicating their detection and interpretation by human observers. For example, \texttt{eva\_giant\_patch14\_336.m30m\_ft\_in22k\_in1k} has the highest object recognition performance among these six ViT-based models, along with the highest perturbation tolerance. However, its adversarial alignment score is negative and anticorrelated with features humans rely on for recognition: the attack primarily affects the background regions instead of the foreground object. The observation suggests that there is a trade-off between perturbation tolerance and adversarial alignment, especially for large-scale and high-accuracy models.

% %%%%%%%%%%%%%%%%%%%%%%%%%%%%%%%%%%%%%%%%%%%%%%%%%%%%%%%%%%%%
\paragraph{Results of $\ell_{\infty}$ PGD Attack.} \label{si_sec:linf_pgd}

\begin{figure}[t!]
\begin{center}
   \includegraphics[width=1\linewidth]{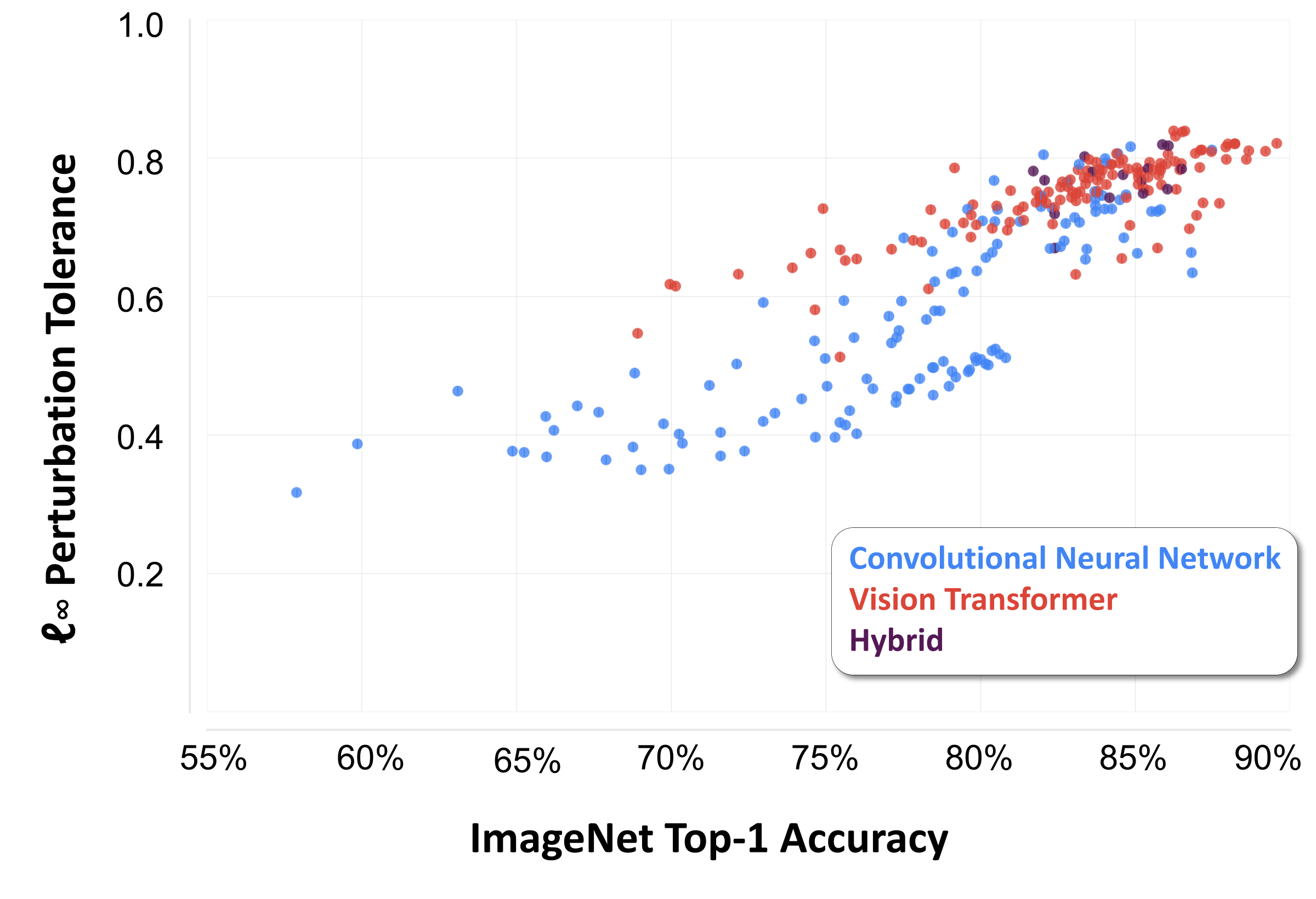}
\end{center}
   \caption{\textbf{The perturbation tolerance of DNNs based on $\ell_\infty$ PGD attack increases as they have improved on ImageNet.}}
\label{si_fig:SI_PGD_Linf}
\end{figure}

Our earlier findings highlight the valuable protection that the DNN scale offers against adversarial attacks. Notably, we also found this finding holds true when considering the $\ell_\infty$ PGD attack, as can been seen from Fig.~\ref{si_fig:SI_PGD_Linf}. This translation of results to $\ell_\infty$ PGD attacks (correlation between $\ell_2$ and $\ell_\infty$ perturbation tolerance: $\rho_{s} = 0.72, p < 0.001$, Fig.~\ref{si_fig:SI_PGD_Linf}) provides additional evidence supporting the promising impact of optimizing DNNs for ImageNet performance in building robust models that can withstand image perturbations. This further strengthens the idea that DNNs with higher accuracy on ImageNet tend to display increased resilience against adversarial perturbations.

%%%%%%%%%%%%%%%%%%%%%%%%%%%%%%%%%%%%%%%%%%%%%%%%%%%%%%%%%%%%

\paragraph{$\ell_2$ Carlini-Wagner \& $\ell_2$ PGD Attacks.}

\begin{figure}[t!]
\begin{center}
   \includegraphics[width=1\linewidth]{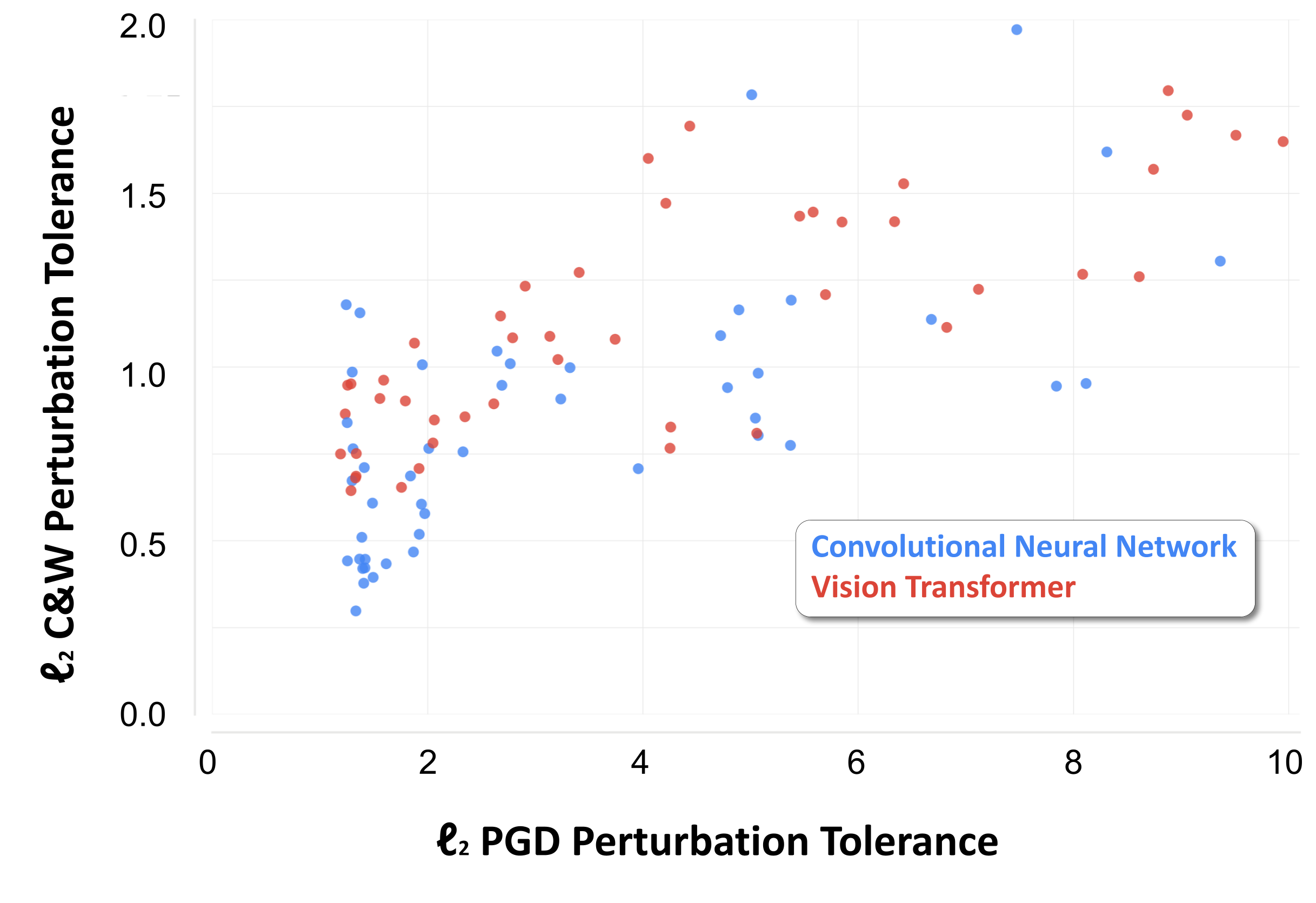}
\end{center}
   \caption{\textbf{A comparison between $\ell_2$ PGD attack and $\ell_2$ C\&W attack on perturbation tolerance.}}
\label{si_fig:SI_PGD_CW_L2}
\end{figure}

\begin{figure}[t!]
\begin{center}
   \includegraphics[width=1\linewidth]{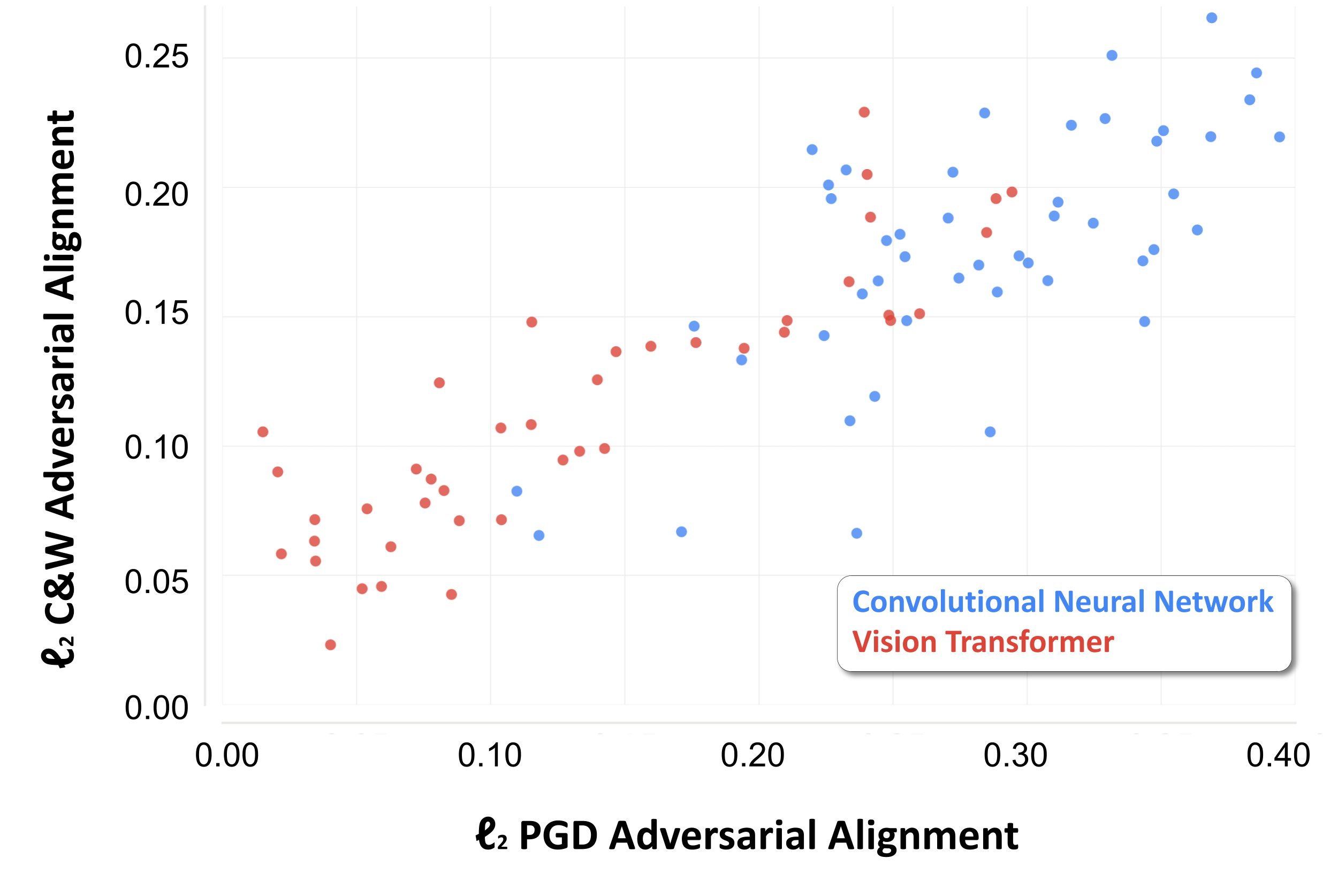}
\end{center}
   \caption{\textbf{A comparison between $\ell_2$ PGD attack and $\ell_2$ C\&W attack on adversarial alignment.}}
\label{si_fig:SI_PGD_CW_spearman}
\end{figure}

Despite the power of the Carlini-Wagner (C\&W) attack, it is known for being extremely slow to compute, which not only requires more gradient steps than PGD but also requires the tuning of an extra parameter denoted $c$. To understand if they are correlated with the PGD attacks we relied on throughout this work, we ran a small-scale survey of DNN tolerance to CW attacks, involving 50 CNNs and 50 ViTs from our model zoo, and 100 images from our 1000 image stimulus set. We found a similar pattern of results with CW as we did with the $\ell_2$ PGD attack (perturbation tolerance: $\rho_{s} = 0.71, p < 0.001$, Fig.~\ref{si_fig:SI_PGD_CW_L2}; adversarial alignment: $\rho_{s} = 0.87, p < 0.001$, Fig.~\ref{si_fig:SI_PGD_CW_spearman}).

\end{document}